\pdfoutput=1

\documentclass[11pt]{article}

\usepackage{acl}

\usepackage{times}
\usepackage{latexsym}

\usepackage[T1]{fontenc}

\usepackage[utf8]{inputenc}

\usepackage{microtype}

\usepackage{bm}
\usepackage{bbm}
\usepackage{amsmath}
\usepackage{booktabs}
\usepackage{tabularx}
\usepackage{graphicx}
\usepackage{multirow}
\usepackage{tablefootnote}
\usepackage{subcaption} %
\usepackage{listings}
\usepackage{enumitem}

\usepackage[normalem]{ulem}

\newcommand{\PreserveBackslash}[1]{\let\temp=\\#1\let\\=\temp}
\newcolumntype{C}[1]{>{\PreserveBackslash\centering}p{#1}}
\newcolumntype{R}[1]{>{\PreserveBackslash\raggedleft}p{#1}}
\newcolumntype{L}[1]{>{\PreserveBackslash\raggedright}p{#1}}

\newcommand{\samtype}[1]{\textsc{#1}}

\usepackage{array}
\newcolumntype{C}[1]{>{\centering\arraybackslash}p{#1}}

\title{Full-Text Argumentation Mining on Scientific Publications}

\author{Arne Binder\textsuperscript{\textnormal{1}} ~~~ Bhuvanesh Verma\textsuperscript{\textnormal{2}} ~~~ Leonhard Hennig\textsuperscript{\textnormal{1}} \\
\textsuperscript{1}German Research Center for Artificial Intelligence (DFKI)  \\ 
\textsuperscript{2}University of Potsdam \\
\textsuperscript{1}{\{\textit{arne.binder, leonhard.hennig}\}\textit{@dfki.de}} \\
\textsuperscript{2}\textit{bhuvanesh.verma@uni-potsdam.de}}

\date{October 2022}

\begin{document}
\maketitle

\begin{abstract}
Scholarly Argumentation Mining (SAM) has recently gained attention due to its potential to help scholars with the rapid growth of published scientific literature. It comprises two subtasks: argumentative discourse unit recognition (ADUR) and argumentative relation extraction (ARE), both of which are challenging since they require e.g.\ the integration of domain knowledge, the detection of implicit statements, and the disambiguation of argument structure~\cite{al_khatib_argument_2021}. While previous work focused on dataset construction and baseline methods for specific document sections, such as abstract or results, full-text scholarly argumentation mining has seen little progress. In this work, we introduce a sequential pipeline model combining ADUR and ARE for full-text SAM, and provide a first analysis of the performance of pretrained language models (PLMs) on both subtasks. We establish a new SotA for ADUR on the Sci-Arg corpus, outperforming the previous best reported result by a large margin (+7\% F1). We also present the first results for ARE, and thus for the full AM pipeline, on this benchmark dataset. 
Our detailed error analysis reveals that non-contiguous ADUs as well as the interpretation of discourse connectors pose major challenges and that data annotation needs to be more consistent.

\end{abstract}

\section{Introduction}

Argumentation Mining (AM) is concerned with the detection of the argumentative structure of text \cite{stede_argumentation_2018}. It is commonly organized into two subtasks: 1) Recognition of argumentative discourse units (ADUs), i.e.\ detecting argumentative spans of text and classifying them into types such as \textit{claim} or \textit{premise}, and 2) determining which ADUs have a relationship to each other and of what kind, e.g.\ \textit{support} or \textit{attack}. Consider the following example, where the premise $P$ supports the claim $C$:
\newpage
\begin{quote}
    \uwave{Dot-product attention is much faster than additive attention}$_\text{C}$, since \uline{it can be implemented using highly optimized matrix multiplication code}$_\text{P}$.\footnote{replicated from \citet{vaswani_attention_2017}}
\end{quote}

Since the amount of published scientific literature is growing exponentially \cite{fortunato_science_2018}, there is recently an increased interest in scholarly argumentation mining (SAM). Understanding the argumentative structure is key, not just to efficiently digest such work, but also to assess its quality \cite{walton_informal_2001}. %
Solving scholarly AM is challenging, because it requires, among other things, the use of domain knowledge, the detection of implicit statements, and the disambiguation of argument structure \cite{al_khatib_argument_2021}. This is even harder when handling full-text that is often less concise and standardized, than, for example, abstracts. 

Previous work in SAM has focused on dataset construction \cite{teufel-moens-1999-discourse,lauscher-etal-2018-argument}, ADU recognition \cite{lauscher-etal-2018-arguminsci,li-etal-2021-scientific}, and the analysis of specific document sections, such as abstract or results \cite{dasigiExperimentSegmentationScientific2017,accuosto-saggion-2019-transferring,mayerTransformerbasedArgumentMining2020}. 
However, to get a thorough understanding of a scientific publication, all parts of the document matter. Ideally, they back up the main argumentation and usually contain details that are relevant for the knowledgeable reader, thus, they should not be neglected. However, since the task is very complex, also for humans, there is not much training data for full-text SAM available.

\begin{figure*}[!ht]
  \includegraphics[width=\textwidth]{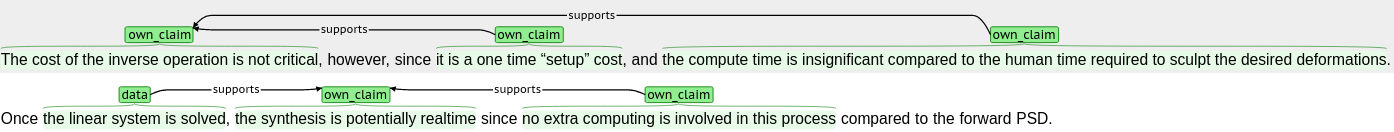}
  \caption{Example with argumentative structure from the Sci-Arg dataset.}
  \label{fig:data}
\end{figure*}

Pretrained Language Models (PLMs) such as SciBERT~\cite{beltagy-etal-2019-scibert} may help to address the above challenges because they contain a lot of linguistic and domain knowledge and have better long-range capabilities, allowing for improved contextualisation, especially when training data is rare. 
We hence propose a PLM based model for full-text SAM. To summarize, our contributions in this work are: 
\begin{itemize}
  \item We are the first to investigate PLMs for full-text SAM, and to present a sequential pipeline for both ADU recognition and argumentative RE on full-text scientific publications (Section~\ref{sec:models}).
  \item Our experimental results show that a SciBERT-based ADU recognition model improves over the state-of-the-art by $+7\%$ F1-score. %
  We present the first relation extraction baseline for the Sci-Args corpus and achieve strong $0.74$ F1 (Section~\ref{subsec:results}).
  \item Our detailed error analysis reveals open challenges and possible ways of improvements
  (Section~\ref{subsec:error_analysis}).
\end{itemize}

\section{Preliminaries}
\label{sec:preliminaries}

We first define the two tasks of ADUR and ARE, and discuss differences to the standard Information Extraction (IE) tasks of Named Entity Recognition (NER) and Relation Extraction (RE).

An Argumentative Discourse Unit (ADU) can be defined as ``span of text that plays a single role for an argument being analyzed and is demarcated by neighboring text spans that play a different role, or none at all'' \cite{stede_argumentation_2018}. It is the smallest unit of argumentation, and  may span anything from an in-sentence clause up to multiple full sentences. ADU recognition requires both detecting argumentative spans, as well as classifying them into predefined categories. Typically, this is realised as sequence tagging task similar to NER, where a sequence of tokens $X = \{t_1, t_2, ..., t_N\}$ is assigned with a corresponding $N$-length sequence of labels $Y = \{l_1, l_2, ..., l_N\}$ with $l_i \in C$ where $C$ is the set of tags that result from converting the ADU types into a tagging scheme like BIO2.\footnote{BIO2: \textbf{B}egin, \textbf{I}nside, \textbf{O}utside of an entity} %
In scholarly AM, common ADU classes are \textit{(Own / Background) Claim}, and \textit{Evidence}, \textit{Data}, or \textit{Warrant} \cite{green-2014-towards,lauscher-etal-2018-argument}. 

In contrast to NER, ADUs typically vary much more in length than named entities. They are also highly context dependent and often discontinuous. ADUR is also related to discourse segmentation, but depends more on broader context and semantics instead of linguistic structure. Elementary Discourse Units (EDUs), the building blocks in the context of Rethorical Structure Theory \cite{mannRhetoricalStructureTheory1988}, are more fine-grained, of shorter length and usually cover the complete text which is less the case for argumentative units. %

Argumentative Relation Extraction is usually defined as classifying a pair of ADUs, \textit{head} and \textit{tail}, as either an instance of one of the target types or the artificial \samtype{no-relation} type. In other words, the task is to assign a label $Y \in C \cup \{\text{\samtype{no-relation}}\}$ to a given input $X = \{T, h, t\}$, where $C$ is the set of relation types, $T$ is the text and $h = (s_h, e_h, l_h)$ and $t = (s_t, e_t, l_t)$ describe the candidate head and tail entities where $s$ and $e$ are the start and end indices with respect to $T$ and $l$ is the entity type. %
Typical relation types for SAM are \textit{Supports}, \textit{Mentions}, \textit{Attacks}, \textit{Contradicts}, and \textit{Contrasts} \cite{lauscher-etal-2018-argument,accuosto-saggion-2019-transferring,nicholsonSciteSmartCitation2021}.

ARE is very similar to standard RE, but SAM relations are often marked by syntactic cues such as connectors, e.g.\ ``because'', ``however'', or ``but'', whereas in common RE, content words like verbs and nouns are typical relation triggers. This makes ARE challenging because these connectors do not always realise argumentative structure, but also mark other aspects of discourse.
Consider, for example, the different meanings of ``while'' in the following example:
\begin{enumerate}[noitemsep]
    \item \uline{While} I love a romantic dinner, I also like fast food.
    \item \uwave{While} I prepare dinner, I watch a movie.
\end{enumerate}
Here, the ``while'' in sentence 1) has a contrastive meaning, whereas sentence 2) denotes a temporal aspect. %

\section{Models}
\label{sec:models}
\begin{figure}[t!]
  \begin{subfigure}[c]{0.48\textwidth}
    \includegraphics[width=\textwidth]{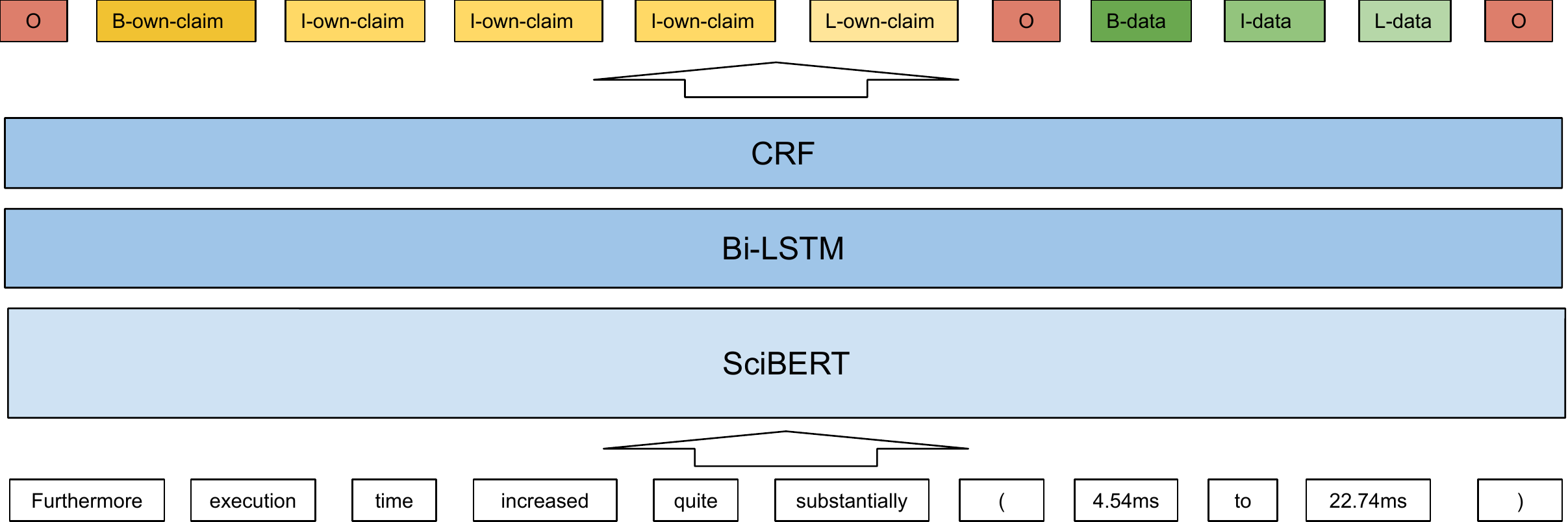}
    \caption{\textbf{ADU Recognition.} Tokens are embedded with a frozen PLM, further contextualized with a trained LSTM followed by a CRF to calculate the tag sequence.}
    \label{fig:model_adu}
  \end{subfigure}
  \vfill
  \vspace{5mm}
  \begin{subfigure}[c]{0.48\textwidth}
    \includegraphics[width=\textwidth]{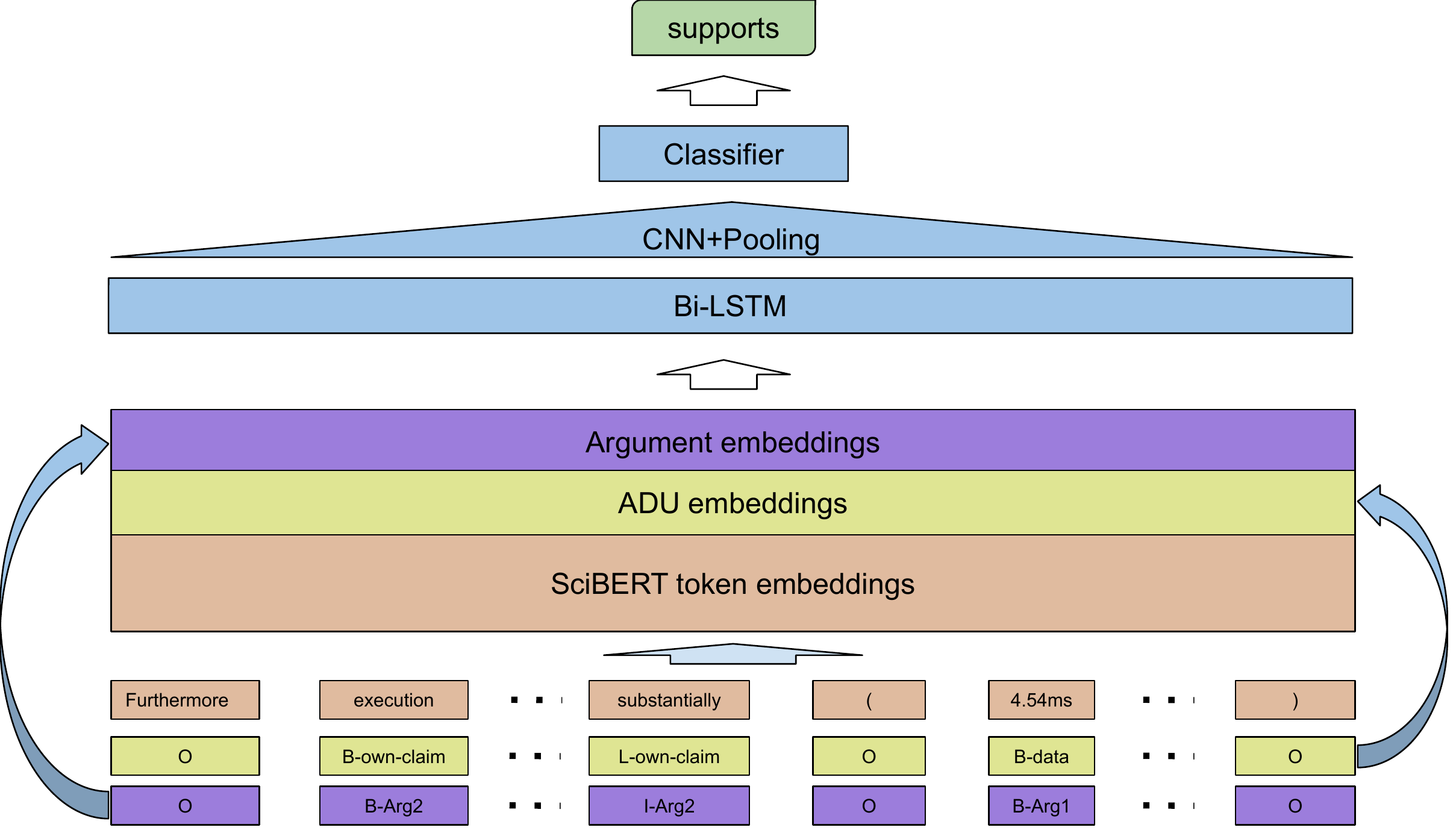}
    \caption{\textbf{Argumentative RE.} Tokens are embedded with a frozen PLM, ADU tags and argument tags are embedded with simple embedding matrices. Embeddings are concatenated, contextualized with a LSTM and converted into a single vector that gets classified by a single fully connected layer.}
    \label{fig:model_rel}
  \end{subfigure}
  \caption{Model setup for (a) ADUR (top) and (b) ARE (bottom).}
  \label{fig:models}
\end{figure}
We propose a pipeline of two distinct models, one for each subtask, that are described in the following.

\paragraph{ADU Recognition (ADUR).} 
The architecture of the ADUR model is visualized in Figure~\ref{fig:model_adu}. We first %
embed the token sequence with a frozen PLM encoder. For sequences that exceed the maximum input length of the embedding model, we process the sequence piece-wise and concatenate the result afterwards. %
The embedded tokens are then fed into a BiLSTM \cite{schuster_bidirectional_1997}. Finally, a Conditional Random Field (CRF) \cite{lafferty_conditional_2001} is used to obtain the label probabilities for each token. We use a combination of a frozen PLM with a trainable contextualization (LSTM) on top because its training requires less resources than fine-tuning the PLM and initial tests have shown similar performance.\footnote{Note that the training dataset is relative small, so restricting the number of trainable parameters seems to mitigate overfitting.}

\paragraph{Argumentative RE (ARE).} The model architecture for the relation extraction subtask is shown in Figure~\ref{fig:model_rel}. ARE is implemented as a classification task, where a pair of candidate ADUs is selected and marked in the input token sequence. To reduce combinatorial complexity, only ADU pairs with a distance smaller than some threshold \texttt{d} are considered. %
Similar to ADU recognition, we first embed the token sequence in a window of \texttt{k} tokens around the candidate entity pair with a frozen PLM model. %
We also create non-contextualized embeddings for the ADU- and argument-tags of the tokens within the window. As argument tags we simply use \textit{head} and \textit{tail} labels to mark the candidate entity tokens. All three embedding sequences are concatenated token-wise  and fed into a BiLSTM. The result is converted into a single vector using a Convolutional Neural Network (CNN)  %
and max-pooling, which then is classified as one of the relation labels by a linear projection with softmax.

\begin{table}[t!]
\centering
\begin{tabular}{l|rr|r}
\toprule
          & \textbf{Train} & \textbf{Test} & \textbf{Total} \\ 
          \midrule \midrule
\textbf{ADUs}                                   &                                        &                                    &                                    \\ \hline
background claim  & 2563                                   & 661          & 3224                               \\
own claim         & 4608                                   & 1241          & 5849                               \\
data               & 3346                                   & 858          & 4204                               \\ 
\midrule \midrule
\textbf{Relations}                               &                                        &                                    &                                    \\ \hline
supports           & 4426                                   & 1260         & 5686                               \\
contradicts        & 551                                    & 133           & 684                                \\
semantically same & 36                                     & 3            & 39                                 \\
\textit{parts of same}    & 1000                                   & 269          & 1269        \\
\bottomrule         

\end{tabular}
\caption{Label counts for the Sci-Arg dataset.}
\label{tab:label_counts}
\end{table}

\section{Experimental Setup}

\paragraph{Dataset.} We use the Sci-Arg dataset \cite{lauscher-etal-2018-argument} for model training and evaluation. It is the only available full text argumentation mining dataset for scientific publications. It contains 40 full text publications annotated with ADUs and argumentative relations. Figure~\ref{fig:data} shows an example excerpt, and Table~\ref{tab:label_counts} summarizes the main dataset statistics. The \samtype{parts of same} relation type is used to model non-contiguous spans. The label counts differ slightly from values published in \citet{lauscher-etal-2018-argument}, because annotations in one file (A28) caused parsing errors and were excluded. Furthermore, non-contiguous spans are not merged. We create a train/test split by using the first 30 documents for training and the remaining 9 for evaluation. 

\paragraph{Preprocessing.} We preprocess the documents by removing the initial XML headers. To decrease the sequence length of the input, we also split the documents into sections, e.g. \textit{introduction} or \textit{conclusion}. This is important to lower computational resource consumption since recent PLMs like SciBERT \cite{beltagy-etal-2019-scibert} usually scale quadratic with the input length and are restricted to a certain max input size, e.g. 512 tokens.
Unfortunately, this leads to the removal of all relations labeled with \samtype{semantically same}, since these connect ADUs from different sections. %
However, this affects only $0.6\%$ of the argumentative relations instances. %

\begin{table}[t]
\centering
\begin{tabular}{lccc}
\toprule
\multicolumn{1}{c}{\textbf{system}} & \multicolumn{2}{c}{\textbf{span based}} & \multicolumn{1}{c}{\textbf{token based}} \\
                                     & exact               & weak               &                                          \\ 
\midrule \midrule
Lauscher 2018c                       &  -                  & -                  & 0.447                                     \\
ours                                 & 0.532                 & 0.668                & 0.518                                     \\ 
\midrule
human                                & 0.602                & 0.729               &                           -              \\
\bottomrule
\end{tabular}
\caption{\textbf{ADU Recognition Performance} as F1 macro average over classes. For \textit{weak} metrics, the gold and the predicted span have to match for at least the half of the characters of the longer span.}
\label{tab:result_adu}
\end{table}

\paragraph{Data Augmentation.}
If the pair of ADUs $(A, B)$ is part of an argumentative relation, it is wrong to assume that $B$ is argumentatively unrelated with $A$, i.e. $(B, A)$ should not be in the \samtype{no relation} class. Thus, we add reversed instances for each available relation in the dataset with the special label \samtype{supports rev} in the case of \samtype{supports} and keep the labels for \samtype{contradicts} and \samtype{parts of same} since these relations are symmetric. In addition to the positive training instances, we also sample negative relation instances from all possible ADU pairs that are no instances of any argumentative relation.%

\paragraph{Training Objective.} We use the the cross entropy loss \cite{rubinstein_cross-entropy_1999} as the training objective for both models \(f_{ADU}\) and \(f_{RE}\):
\begin{equation*}
\begin{split}
\mathcal L_{CE}&(y, \hat{y}; {\bm\theta}):= 
-f_{\bm\theta}(y) \cdot \log f_{\bm\theta}(\hat{y})
\end{split}
\end{equation*}
where \(y\) and \(\hat{y}\) are the target and predicted probabilities for the token or relation labels, respectively, and \({\bm\theta}\) is the set of trainable model parameters. In the case of ADU recognition, we obtain the best tagging sequence via Viterbi Decoding \cite{viterbi_error_1967}, as usual for CRF-based models.

\begin{table}[ht]
\centering
\begin{tabular}{lcc}
\toprule
    &      F1-exact & F1-weak \\
\midrule
\midrule
@gold ADUs & \multicolumn{2}{c}{0.739} \\
@predicted ADUs  & 	0.210 &   0.310\\
\midrule
human &  0.341 & 0.469 \\
\bottomrule
\end{tabular}
\caption{\textbf{Argumentative RE Performance} as micro average over classes with provided gold ADUs (@gold ADUS) or ADUs predicted with our entity recognition model (@predicted ADUs), i.e. the full relation extraction pipeline. \textit{human} indicates inter-annotator-agreement for the corpus data \cite{lauscher-etal-2018-investigating} which is comparable to \textit{@predicted ADUs}. For \textit{weak} metrics, best weakly matching ADUs are calculated first, then predicted relations are mapped to these and finally metrics are calculated as usual. }
\end{table}

\paragraph{Metrics.} Since we compare against evaluation results from \citet{lauscher_investigating_2018}, we adopt their metrics for ADU recognition, namely a token-based F1-score that is macro-averaged over classes. However, we also compute \textit{span-based} macro-F1 scores in two variants as described in \citet{lauscher-etal-2018-argument}: For \textit{exact} span-based metrics, the recognized ADU has to match exactly for start and end indices, as well as ADU type. For \textit{weak} matches, the ADU has to match in type, but the target and predicted spans only have to overlap by at least the half of the length of the shorter span. %
Weak match evaluation is motivated by considerable length and variance of ADU expressions, which makes exact matches difficult, and also allows for comparison with human annotator agreement scores as presented in \citet{lauscher-etal-2018-investigating}. 

For the relation recognition task, we follow the literature and present micro-averaged F1 scores. Similar to ADU recognition metrics, we calculate weak metrics by first determining target ADUs that can be assigned to predicted ADUs in the way of weak ADU matching as described above, and then calculate F1 scores as usual~\cite{lauscher-etal-2018-argument}. Note that \samtype{parts of same} is just a helper relation, so we merge ADUs connected by this relation type first, and then compute scores over the remaining relation types.     

\paragraph{Training Details \& Hyperparameters.} For both tasks, we first conduct a hyperparameter search. We use token-based macro-F1 as the optimization target for ADU recognition and micro-F1 as the target for relation classification. Final hyperparameter values are listed in Appendix~\ref{sebsec:appendix_experimenta_setup_and_hyperparameters}.

Since there is no dev split, we perform 5-fold cross validation for each subtask on the train split with the best hyperparameter settings and different random seeds for parameter initialization.  The best of these 5 models are used for the final evaluation. Detailed training configurations, logs and statistics for the ADU recognition and the ARE subtasks are collected within the Weights \& Biases framework.\footnote{see \url{https://wandb.ai}} We make these and our source code publicly available for better reproducibility of our experiments.\footnote{For ADU recognition, see \url{https://wandb.ai/sam_dfki/best_adu_uncased}, for argumentative RE, see \url{https://wandb.ai/sam_dfki/best_rel_uncased}, and for the source code, see \url{https://github.com/DFKI-NLP/sam}.} %

\begin{table}
\centering
\begin{tabular}{llccc}
\toprule
       &           &  P &  R &   F1 \\
\midrule
\midrule
       & background claim &       0.56 &    0.44 & 0.49 \\
exact  & own claim &       0.48 &    0.55 & 0.51 \\
       & data &       0.57 &    0.62 & 0.60 \\
       \midrule
       & background claim &       0.77 &    0.60 & 0.68 \\
weak  & own claim &       0.63 &    0.73 & 0.67 \\
       & data &       0.62 &    0.69 & 0.65 \\
           
\bottomrule
\end{tabular}
\caption{\textbf{ADUR Performance per Class.} Macro averaged precision (P), recall (R), and F1.}
\label{tab:result_adu_per_class}
\end{table}

\section{Results and Discussion}

\label{subsec:results_and_discussion}
This section presents our experimental results. First, we compare against the ADU recognition baseline as provided by \citet{lauscher-etal-2018-investigating}. Then, we present findings about prominent error cases and close with an ablation study.

\subsection{Results}
\label{subsec:results}
Table~\ref{tab:result_adu} presents the macro-F1 scores of the ADUR baseline, our approach, and human performance in terms of inter-annotator agreement, as reported in \citet{lauscher-etal-2018-investigating}. Our model achieves $0.518$ token-based F1, significantly outperforming the baseline by $7\%$. The gap to the human performance is also narrow, especially when looking at the weak metrics with relaxed boundary constraints, where our model achieves $92\%$ of to the human score. For exact metrics, the model reaches only $88\%$ of the human performance, suggesting that exact ADU boundary detection is more challenging. The performance of the model for argumentative RE is a strong $0.739$ micro-F1. Note, that we need to merge non-contiguous ADUs first before calculating the ARE scores. We do this via predicted \samtype{parts of same} relations, which are recognized with a F1-score of $0.860$. For the full pipeline, the model achieves a respectable $0.210$ micro-F1 score, which corresponds to $62\%$ of the human performance.  

\subsection{Error Analysis}
\label{subsec:error_analysis}
\paragraph{ADUR Error Analysis.} The decrease in performance when comparing weak with exact metrics is high for the classes \samtype{background claim} ($-28\%$) and \samtype{own claim} ($-24\%$), but low for class \samtype{data} ($-8\%$), see Figure~\ref{tab:result_adu_per_class}. This may be because the latter is mainly about references or mentions of concise facts where boundaries are much easier to detect. 

\begin{table}[t!]
\centering
\begin{tabular}{lccc}
\toprule
    &  P &  R &   F1 \\
\midrule
\midrule
contradicts &       0.505 &    0.724 & 0.595 \\
supports &       0.739 &    0.774 & 0.756 \\
\bottomrule
\end{tabular}
\caption{\textbf{ARE Performance per Class}. Micro averaged precision (P), recall (R), and F1 on gold ADUs. Note that non contiguous ADUs linked via predicted \samtype{parts of same} relations are merged first before calculating the scores.}
\label{tab:result_rel_per_class}
\end{table}

\begin{figure*}[ht]
    
    \hspace*{5mm}%
    \begin{subfigure}[c]{0.43\textwidth}
        \includegraphics[width=\textwidth]{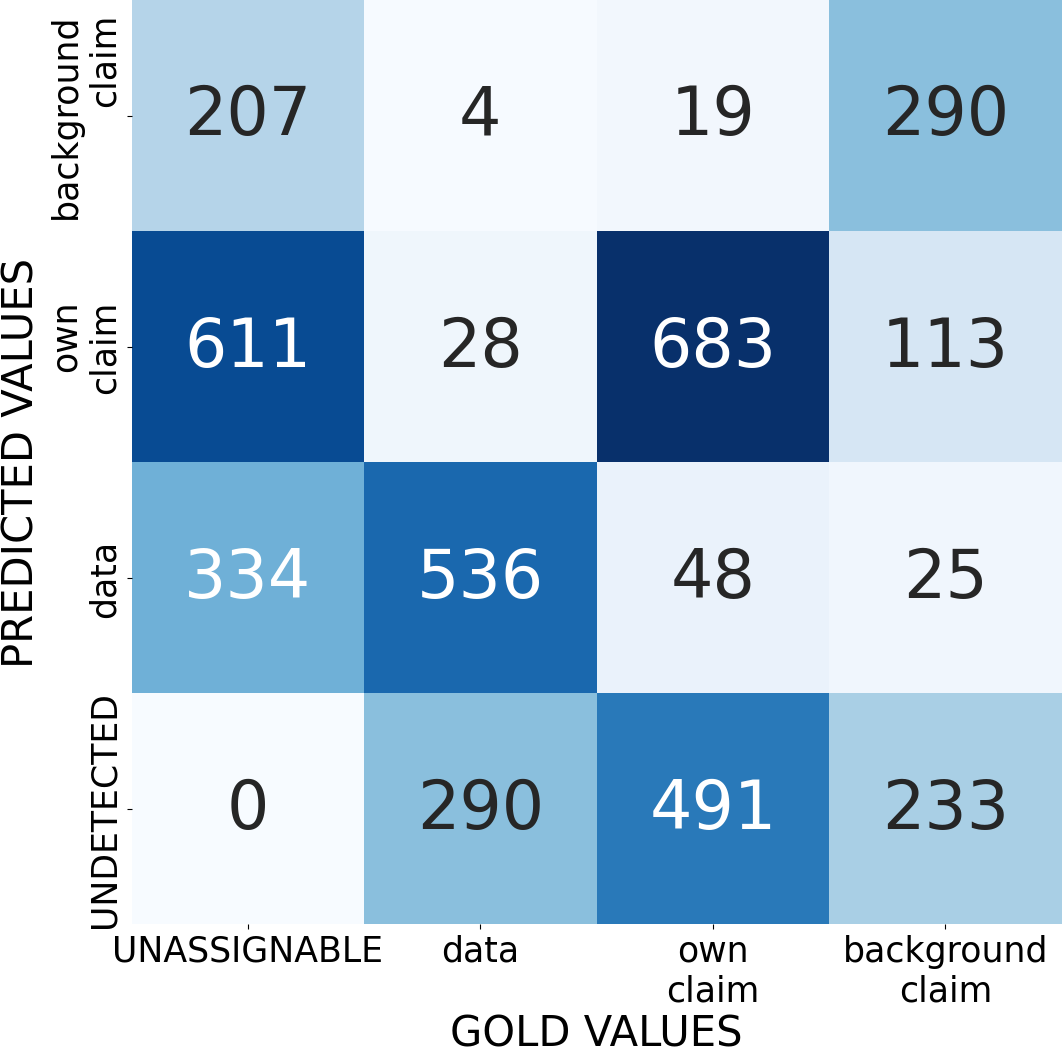}
    \end{subfigure}
    \hfill%
    \begin{subfigure}[c]{0.42\textwidth}
        \includegraphics[width=\textwidth]{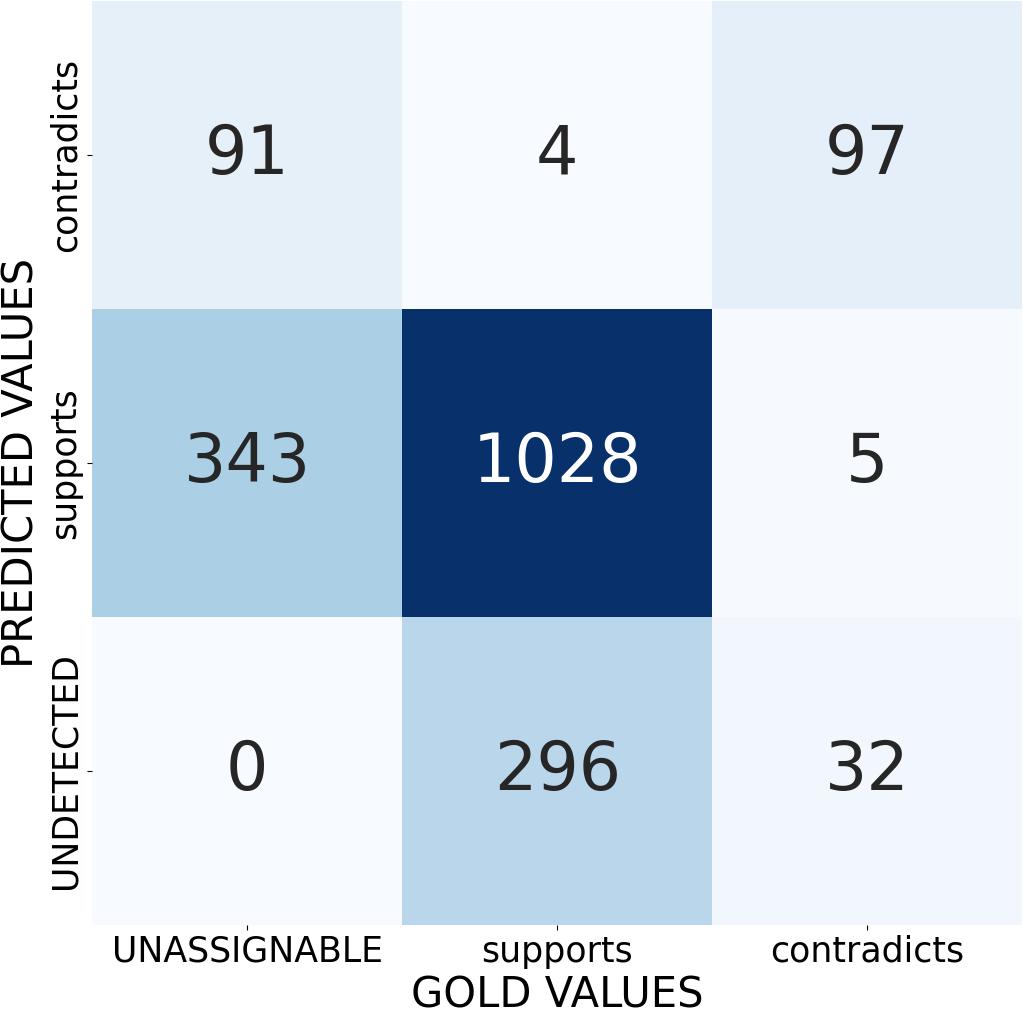}
    \end{subfigure}%
    \hspace*{\fill}%
    \caption{Confusion matrices for ADU recognition (left) and argumentative RE (right).}
    \label{fig:confusion_both}
\end{figure*}

Most of the errors originate from \textit{detecting} ADUs, i.e.\ deciding if a text span is an ADU from any type, in comparison to \textit{classifying} a detected ADU span into one of types. The exact span-based macro-F1 for the subtask of ADU \textit{classification} is $0.854$, whereas the respective score for ADU \textit{detection} is only $0.617$. This difference is even larger for the RE subtask where the micro-F1 is $0.749$ for relation \textit{detection} and $0.992$ (!) for relation \textit{classification}. Figure~\ref{fig:confusion_both} shows the confusion matrices for the ADUR and ARE subtasks.

Interestingly, many of ADU classification errors ($48\%$) are instances of type \samtype{background claim}, where the model predicts \samtype{own claim} instead, indicated by low precision for \samtype{own claim} and low recall for \samtype{background claim} as shown in Table~\ref{tab:result_adu_per_class}. %
Looking into these misclassifications revealed the following main challenges (in order of decreasing frequency): 1) an island of one or two background claims surrounded by many own claims or located at the border between regions of these two types, 2) the ADU is linked via the structural \samtype{parts of same} relation, i.e. it is split by some other content and at least one part of the complete ADU is not detected correctly, and 3) mentions of the author in a background claim (e.g. "[A] drawback of this model for \textit{our} application is [...]" or "It enables \textit{us} to model [...]"). Issues 1) and 2) may suggest that looking at the sequence of ADU types or linguistic surface features is not enough and a deeper "understanding" and/or domain knowledge are required, especially since the training data is very limited. \citet{lauscher-etal-2018-investigating,accuosto-saggion-2019-transferring} analyse the impact of SAM to related tasks, suggesting to train on these may mitigate this issue.
Finally, issue 2) may be improved by using a joint ADUR+ARE model or an ADUR model that allows to predict non-contiguous spans. Note that we tackle ADU detection in fact with both models in combination because we require the \samtype{parts of same} predictions to merge the respective ADUs. This poses a challenge for both models: The ADUR model is trained to predict incomplete instances and the ARE model  needs to handle instances from conceptional different types of classes, i.e.\ argumentative and structural relations.

\paragraph{ARE Error Analysis.} For the relation extraction subtask, the general performance is higher than for ADUR with approximately only one third of false negatives or false positives with respect to true positive. However, the performance for \samtype{contradicts} is much lower than for \samtype{supports}, see Figure~\ref{tab:result_rel_per_class}. On reason appears to be the class imbalance. There are substantially less training instances for that class (ratio of $1 : 8$, see Figure~\ref{tab:label_counts}). Furthermore, the model significantly overpredicts the \samtype{contradicts} relations (see confusion matrix in Figure~\ref{fig:confusion_both}). To unravel this phenomenon, we manually analysed 255 relation candidates from different error categories (true positives, false positives, and false negatives). This revealed, that most of the instances falsely predicted as \samtype{contradicts} can be associated with specific linguistic surface features, especially occurrences of discourse connectors like ``however'' that are commonly used to express contrastive ideas, but not in this case (see the example in the end of Section~\ref{sec:preliminaries}). Apparently, the model overfits on these shallow markers which is further supported by the fact that all analysed correctly predicted relation instances of that type could be associated with entries of a small set of connectors.\footnote{Consisting of (in decreasing order of frequency): ``however'', ``but'', ``while'', ``in contrast'', ``though'', ``despite'', and ``even though''.}

Regarding the \samtype{supports} relation, the analysis revealed that sentence boundaries seem to be a very strong signal. An over-proportional amount ($85\%$) of correct predictions has both arguments in the same sentence compared to $20\%$ and $15\%$ for false positives and false negatives, respectively. This is even stronger when taking the argument types into account: \samtype{supports} relations that are in the same sentence and connect a \samtype{data} ADU with any claim ADU make up for $88\%$ true positives, but only for $19\%$ and $12\%$ of false positives and false negatives. Note, that per definition of the Sci-Arg annotation scheme\footnote{The original annotation guidelines can be found here: \url{http://data.dws.informatik.uni-mannheim.de/sci-arg/annotation_guidelines.pdf}} \samtype{data} never participates in a \samtype{contrast} relation which may be one reason why relation classification performance is so high. More detailed results of the manual analysis can be found in Figure~\ref{fig:error_analysis_rel} in the Appendix.

During our analysis we noticed a reasonable amount of potentially mislabeled relation instances ($16\%$), especially missing support relations between \samtype{own claim}s. Table~\ref{tab:mislabeled} shows some examples where relations were correctly detected by the ARE model, but they do not exist in the gold data.

\begin{table*}
\centering
\begin{tabular}{p{0.67\textwidth}|C{0.13\textwidth}|C{0.1\textwidth}}
    \toprule
        Text with ADUs  &  Annotated & Correction \\
    \midrule
    \midrule
        As \uwave{the calculations of the wrinkling coefficients are done on a per triangle basis}$_\text{DATA}^{\bm{A}}$, \uline{the computational time is linear with respect to number of triangles}$_\text{OWN CLAIM}^{\bm{B}}$. & 
        {\centering$A \leftarrow_S B$ } & 
        $A \rightarrow_S B$  \\  %
    \midrule
        \uwave{There are several possibilities to deal with this restriction}$_\text{OWN CLAIM}^{\bm{A}}$. \uline{One could decide to restrict the simulations to small deformations where the approximation is valid}$_\text{OWN CLAIM}^{\bm{B}}$. &
        {\centering - } & 
        $A \leftarrow_S B$ \\  %
    \midrule
        As stated in \uline{Section 3.3}$_\text{DATA}^{\bm{A}}$, \uwave{two different wrinkle patterns give different wrinkling coefficients for the same triangle geometry}$_\text{OWN CLAIM}^{\bm{B}}$. Hence, \uline{for the same deformation of the triangle}$_\text{DATA}^{\bm{C}}$, corresponding to each pattern, \uwave{the modulation factors will be different}$_\text{OWN CLAIM}^{\bm{D}}$. & 
        {\centering $A \rightarrow_S B$ \\ $C \rightarrow_S D$ } &
        {\centering $A \rightarrow_S B$ \\ $C \rightarrow_S D$\\ $B \rightarrow_S D$} \\ %
    
    \midrule
        \uline{If a pattern is orthogonal to the deformation direction}$_\text{OWN CLAIM}^{\bm{A}}$ (as compared to the other), \uwave{corresponding modulation factor will be small}$_\text{OWN CLAIM}^{\bm{B}}$. In other words, \uline{the direction of the deformation favors one pattern over the other}$_\text{OWN CLAIM}^{\bm{C}}$. & 
        {\centering $A \rightarrow_S B$ } & 
        {\centering $A \rightarrow_S B$ \\ $B \rightarrow_S C$} \\ %
           
    \bottomrule
\end{tabular}
\caption{Examples for potentially mislabeled relation instances. $A \rightarrow_S B$ means that the pair of ADUs $(A, B)$ is an instance of the \samtype{supports} relation. All proposed corrections are predicted by our model.}
\label{tab:mislabeled}
\end{table*}

\subsection{Ablation Study}
\label{subsec:ablation_study}
We analysed the effect of our approach to add reversed relations. We trained another set of models in a 5-fold cross validation setting with same hyperparameters, but without the augmentation. The resulting mean bootstrapped %
micro F1 is $0.601$, significantly lower than the mean result %
with augmentation enabled which is $0.762$ with $p < 1\text{e-}10$.
We gather bootstrapped scores by randomly sampling 10 test document sections, calculate the scores for both model variants as usual and repeat that process for 100 times. Note that there are 114 document sections in total after preprocessing the test set.

\section{Related Work}
AM is intensively studied for domains like public debates, essays, or legal texts \cite{lawrence-reed-2019-argument}. As one of the earliest work for the scientific domain, \citet{teufel-moens-1999-discourse} proposed Argumentative Zoning (AZ) where sentences are classified as \samtype{aim}, \samtype{contrast}, \samtype{textual}, \samtype{own}, \samtype{background}, \samtype{basis}, or \samtype{other}. The authors created a corpus of 80 annotated full-text papers. They trained Naive-Bayes (NB) and Support-Vector-Machine (SVM) models with hand crafted features and achieved a performance of 0.442 macro-F1. Later work defines similar concepts like "zone of conceptualization" \cite{liakata-2010-zones} with classes like \samtype{Experiment}, \samtype{Background}, or \samtype{Model}, and trained CRF based models on that \cite{liakataAutomaticRecognitionConceptualization2012} (0.18 to 0.76 F1 depending on classes). \citet{guo-etal-2010-identifying} compares these schemes with abstract section name detection and trains NB and SVM models. \citet{dasigiExperimentSegmentationScientific2017} studied the problem of scientific discourse parsing and annotated the result sections of 75 papers with a seven label taxonomy described in \citet{dewaardVerbFormIndicates2012} like \samtype{goal}, \samtype{fact}, or \samtype{hypothesis}. They use an LSTM based model augmented with Attention \cite{vaswani_attention_2017} to obtain sentence representations and present 0.74 F1 performance. In their follow-up work \cite{li-etal-2021-scientific} they achieve a strong 0.841 F1 by using a combination of transfer learning from discourse annotated abstracts (PubMedRCT, \citet{dernoncourt-lee-2017-pubmed}) and a model consisting of SciBERT, Attention, BiLSTM, and CRF. In that respect, their approach is similar to ours for ADUR, however, they apply their methods only on the results section of a document and detect full sentence ADUs only. In a similar vein, \citet{achakulvisutClaimExtractionBiomedical2020} propose a sentence based claim extraction model consisting of BiLSTM and a CRF that they pre-trained on the PubMedRCT dataset. They achieve a performance of 0.790 F1 on a dataset of 1500 abstracts from the Medline dataset. %
\citet{lauscher-etal-2018-arguminsci} proposes a tool for automatic ADU recognition and other tasks. Their models are trained on the Sci-Arg dataset and consist of pre-trained word embeddings and a BiLSTM for token classification tasks (e.g. ADUR) and an additional Attention mechanism to obtain sentence representations for the other tasks.

All work mentioned above focuses primarily on the detection and classification of argumentative components. \citet{stabArgumentationMiningPersuasive2014a} argues for the need to also analyse argumentative structure, e.g.\ to automate knowledge base population or reasonable validate claims because that requires to link the respective premises. They also highlight that discourse theory and data is not suited out of the box for argumentative analysis because discourse relations do not cover relevant argumentative relation types and connect primarily neighboring elements which does not reflect argumentative structure. However, \citet{accuosto-saggion-2019-transferring} propose to derive argumentative structure information from discourse data. They annotate a subset of 60 abstracts from the SciDTB scientific discourse dataset \cite{yang-li-2018-scidtb} with argumentative units and relations. Then, they train models consisting of a BiLSTM, CRF, contextualized word embeddings (ELMo, \citet{peters-etal-2018-deep}) and an encoder pre-trained on the discourse data. They show that adding the encoder significantly improves the performance up to 0.40 F1 argumentative attachment scores, which subsumes argumentative component and relation recognition. \citet{kirschner-etal-2015-linking} created a new corpus by annotating the introduction and discussion sections of 24 scientific articles. The authors consider two argumentative relations, \samtype{support} and \samtype{attack}, and also two discourse relations, \samtype{detail} and \samtype{sequence} borrowed from RST \cite{mannRhetoricalStructureTheory1988}, annotated on the sentence level. Recently, \citet{mayerTransformerbasedArgumentMining2020} proposed an argumentation mining pipeline for ADUR and ARE on a new dataset. They annotate 500 Medline abstracts with \samtype{claim} and \samtype{evidence} ADUs as well as \samtype{support} and \samtype{attack} relations. The authors trained and analysed the performance of different models consisting of encoders, like word embeddings, contextualized word embeddings and BERT variants, in combination with a Gated Recurrent Unit (GRU) or LSTM and a CRF. They present a strong micro-F1 of up to 0.92 for ADUR and a performance of up to 0.69 for the full pipeline and conclude that Transformers, especially domain specific ones like SciBERT, work best for SAM at Medline abstracts. Note that, similar to our weak measures, they count predictions as true positive when $75\%$ of the tokens\footnote{This differs from our weak measures in two ways: Following \citet{lauscher-etal-2018-argument}, we require $50\%$ overlap in means of characters, not tokens.} overlap. Another work \cite{fergadis-etal-2021-argumentation} that analyses the performance of Transformers for SAM proposes a new corpus of 1000 abstracts with sentence level annotations for \samtype{claim} and \samtype{evidence}. The authors use a SciBERT applied sentence wise with a BiLSTM over the CLS token embeddings as contextualizer and present a 0.624 macro-F1.

\section{Conclusion and Future Work}

In this paper, we presented a pipeline based approach to handle full-text argumentation mining on scientific publications and showed its effectiveness by establishing new state-of-the-art performance on the Sci-Arg corpus. However, there is still a significant gap to human performance. We used PLM based models for both subtasks, argumentative discourse unit recognition (ADUR) and argumentative relation extraction (ARE), and found similar improvements gains (+7\%) as reported elsewhere when using Transformers over traditional approaches without Attention mechanism, even without fine-tuning the PLMs. 

Our detailed error analysis revealed several findings. First, recognizing instances is much harder than assigning the correct label, which is true for both tasks, but especially for ARE. The performance suffers from shallow processing, i.e. the models are tricked by linguistic surface features like author referencing pronouns in background claims or non-argumentative discourse connectors. Furthermore, ADUR detection struggles a lot in the context of non-contiguous elements which is reasonable because it is trained with incomplete information. This calls for conceptional better modeling of the task, for instance with a joined model for ADUR and ARE. Finally, we could confirm that SAM is a complex problem that is even hard for humans. However, the low inter-annotator-agreement reported by the Sci-Arg authors and our finding that a significant amount ($16\%$) of the manually analysed ARE instances are questionable labeled raises the need for even more annotation rounds, maybe with multiple domain experts, or a simplified annotation scheme. 
\section*{Acknowledgments}
We would like to thank Aleksandra Gabryszak and the anonymous reviewers for their valuable comments and feedback on the paper. This work has been supported %
by the German Federal Ministry of Education and Research as part of the projects CORA4NLP (01IW20010) and Software Campus 2.0 (01IS17043). %

\bibliography{anthology,custom}
\bibliographystyle{acl_natbib}

\appendix

\begin{figure*}[ht]
    \includegraphics[width=\textwidth]{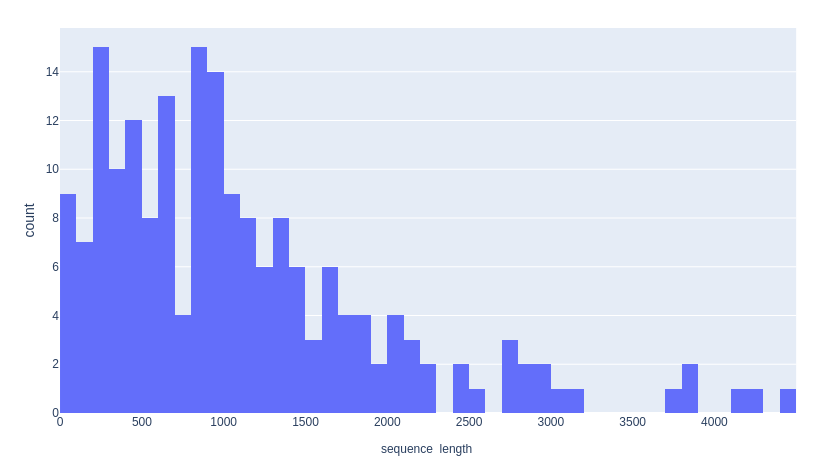}
    \caption{\textbf{Distribution of input sequence lengths.} This is after splitting the document text into sections and tokenization. Note that this is primarily relevant for the ADU model since we use a much smaller token window size $k$ to restrict the input for the ARE model.
    }
    \label{fig:sequence_length_distribution}
\end{figure*}

\section{Appendix}
\subsection{Preprocessing}
\label{subsec:appendix_preprocessing}

We use the following regular expression pattern to match content in the beginning of the files that we remove: ``\texttt{\lstinline[breaklines=true]{<\\?xml[^>]*>[^<]*<Document xmlns:gate=\"http://www.gate.ac.uk\"[^>]*>[^<]*}}''  (without the outer quotes).
Main sections are marked by \texttt{<h1>SECTION\_HEADING</h1>} in the Sci-Arg corpus where \texttt{SECTION\_HEADING} is any text, so we use this regular expression pattern to split the texts: ``\texttt{<H1>}''(without the quotes). Note, that we keep that content in the input. The input sequence lengths for the ADU model reaches still values $>4000$. Figure~\ref{fig:sequence_length_distribution} shows its distribution.

\subsection{Experimental Setup and
Hyperparameters}
\label{sebsec:appendix_experimenta_setup_and_hyperparameters}

We use the AllenNLP framework to implement the models and execute the training. As PLM, we use the uncased variant of SciBERT \cite{beltagy-etal-2019-scibert} as provided by AllenAI\footnote{see \url{https://huggingface.co/allenai/scibert_scivocab_uncased}}. %
ADAM \cite{kingma_adam:_2014} is used as optimizer. We use batch sizes of 8 and 128 for ADU recognition and RE, respectively, that are derived from resource constraints.
The ADU tags are encoded with the BIOUL tagging scheme. For the RE subtask, we hand-picked embedding sizes of 13 and 3 for the ADU-tags and argument-tags, respectively, that are derived form the number of classes.\footnote{Note, the three ADU-tags are each BIOUL encoded and the argument types, \textit{head} and \textit{tail}, are BIO encoded.} 

As a result of the hyperparameter search, we use the following parameters for the ADU recognition task: a learning rate of 0.005, dropout probability of 0.5 before and after the PLM and 0.4394 in the LSTM, a gradient normalization threshold of 7.0, a patience of 20 epochs for early stopping, two layers for the LSTM with a hidden size of 300. In the case of RE, we got the following values: a learning rate of 0.0005, a dropout probability of 0.3061 before and after the PLM and 0.4394 in the LSTM, a gradient normalization threshold of 4.12, 4 layers for the LSTM with a hidden size of 430, 193 filters for the CNN (with ngram sizes of 3, 5, 7, and 10), a hidden size of 860 for the final projection layer, a token window size \texttt{k} of 479 tokens around the center of the candidate argument pair, a max inner token distance \texttt{d} between the arguments of 177\footnote{This causes a loss of 0.23\% of \samtype{support} instances and 0.5\% of \samtype{parts of same} instances, which is neglectable.}, and finally, we use a factor of three for the amount of negative examples, i.e. we add three times as many existing argumentative ADU pairs as \samtype{no relation} instances which we sample from all available pairs without a relation label and within the distance constraint.  

\subsection{Training Resources}
The hyperparameter search was performed on a single Nvidia RTX A6000 (48GB). The training of the final models, i.e. 5 for each subtask, and inference was calculated on single Nvidia GeForce GTX 1080 Ti (12GB). The total training time for all final models was 5h51m for ADUR and 40h17m for ARE.

\begin{figure*}
    \begin{subfigure}[t]{\textwidth}
    
    \centering
        \includegraphics[width=0.9\linewidth]{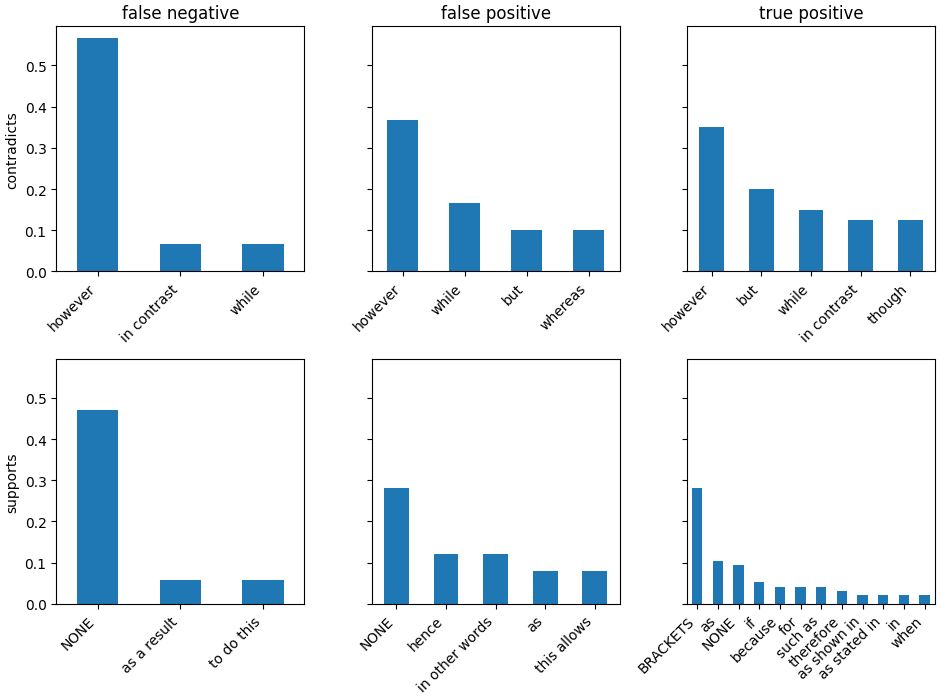}
        \caption{Distribution of \textbf{connecting phrases}. Despite being no real discourse connectors, we also collected markers like BRACKETS that seem to be important surface features. NONE indicates that no connective element was found.}
    \end{subfigure}
    \vfill
    \vspace{5mm}
    \begin{subfigure}[t]{\textwidth}
    \begin{subfigure}[t]{0.47\textwidth}
        \includegraphics[width=\linewidth]{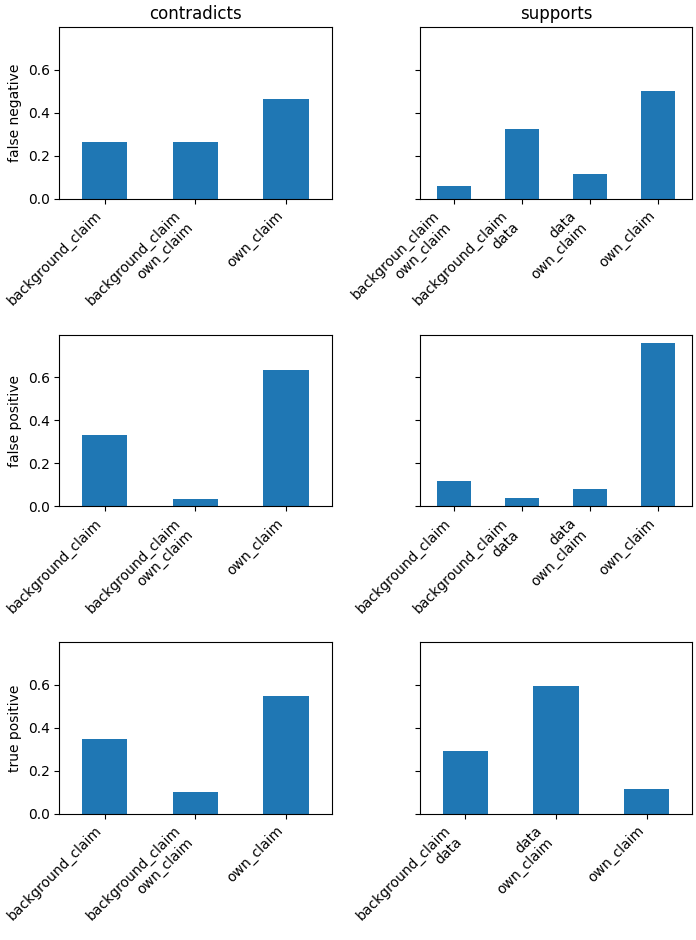}
        \caption{Distribution of \textbf{relation arguments} (sorted and mentioned only once if both arguments are the same).}
    \end{subfigure}
    \hfill
    \begin{subfigure}[t]{0.47\textwidth}
        \includegraphics[width=\linewidth]{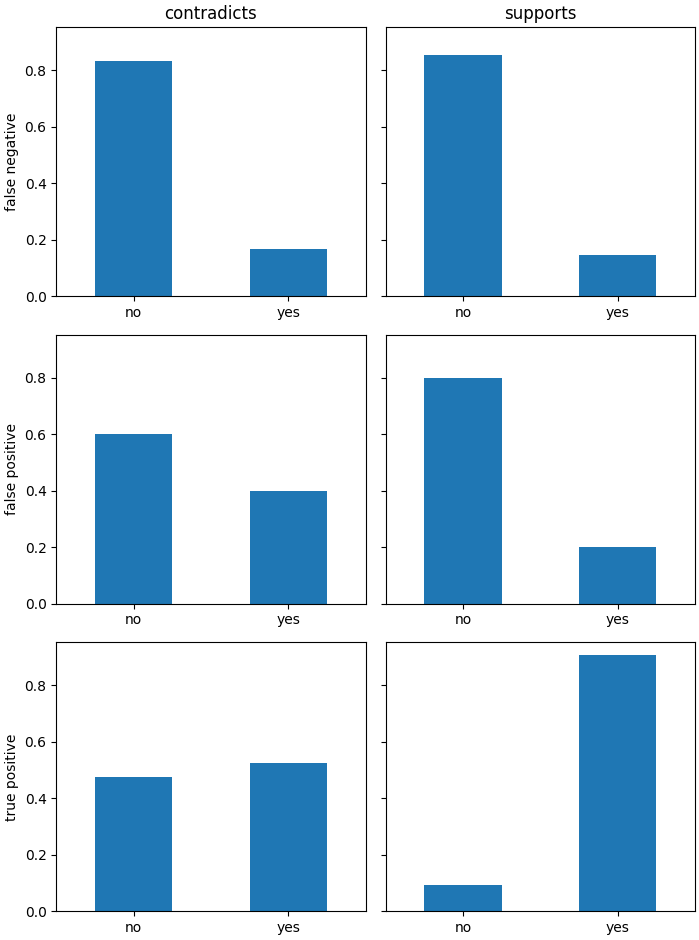}
        \caption{Distribution of the feature that both arguments are \textbf{in the same sentence}.}
    \end{subfigure}
   \end{subfigure}
   \caption{Results of the manual error analysis for argumentative relation extraction. The figures show proportions of different features (connectors, arguments, and same sentence feature) at different subsets by error type (false negative, false positive, or true positive). The lowest entries per category are excluded. Values are calculated on a manually collected subset of 255 relation instances in total. }
   \label{fig:error_analysis_rel}
\end{figure*}

\end{document}